\documentclass[conference]{IEEEtran}
\ifCLASSINFOpdf
\else
\fi
\usepackage{amsmath,amssymb,amsfonts}
\usepackage{cite}
\usepackage{amsmath,amssymb,amsfonts}
\usepackage{algorithmic}
\usepackage{graphicx}
\usepackage{textcomp}
\usepackage{xcolor}
\usepackage{mathtools}
\usepackage{balance}
\usepackage{caption}
\usepackage{subcaption}
\usepackage{enumitem}
\newcommand{\ve}[1]{{\mbox{\boldmath${#1}$}}}

\begin{document}
%
\title{Remaining Useful Life Estimation Using Functional Data Analysis}

\author{\IEEEauthorblockN{Qiyao Wang, Shuai Zheng, Ahmed Farahat, Susumu Serita, Chetan Gupta}
\IEEEauthorblockA{Industrial AI Laboratory, Hitachi America, Ltd. R$\&$D \\
Santa Clara, CA, USA \\
firstname.lastname@hal.hitachi.com}}


%


\maketitle
\footnotetext[1]{978-1-5386-8357-6/19/\$31.00˜\copyright˜2019 IEEE. Personal use of this material is permitted.  Permission from IEEE must be obtained for all other uses, in any current or future media, including reprinting/republishing this material for advertising or promotional purposes, creating new collective works, for resale or redistribution to servers or lists, or reuse of any copyrighted component of this work in other works.}%

\begin{abstract}
Remaining Useful Life (RUL) of an equipment or one of its components is defined as the time left until the equipment or component reaches its end of useful life. Accurate RUL estimation is exceptionally beneficial to Predictive Maintenance, and Prognostics and Health Management (PHM). Data driven approaches which leverage the power of algorithms for RUL estimation using sensor and operational time series data are gaining popularity. Existing algorithms, such as linear regression, Convolutional Neural Network (CNN), Hidden Markov Models (HMMs), and Long Short-Term Memory (LSTM), have their own limitations for the RUL estimation task. In this work, we propose a novel Functional Data Analysis (FDA) method called functional Multilayer Perceptron (functional MLP) for RUL estimation. Functional MLP treats time series data from multiple equipment as a sample of random continuous processes over time. FDA explicitly incorporates both the correlations within the same equipment and the random variations across different equipment's sensor time series into the model. FDA also has the benefit of allowing the relationship between RUL and sensor variables to vary over time. We implement functional MLP on the benchmark NASA C-MAPSS data and evaluate the performance using two popularly-used metrics. Results show the superiority of our algorithm over all the other state-of-the-art methods.  
\end{abstract}


%
\IEEEpeerreviewmaketitle

\section{Introduction}
\label{sec1}
Predictive Maintenance, a vital component in Prognostic and Health Management (PHM), aims to monitor an equipment's condition and designs techniques to actively determine maintenance strategies \cite{mobley2002introduction}. Predictive Maintenance is gaining prominence across different industries, as it effectively increases equipment's uptime, reliability, efficiency, productivity and safety. Researchers have formulated Predictive Maintenance problems from different perspectives, of which the two most important are: \textit{1) remaining useful life estimation}, which aims to estimate the remaining time to the end of equipment's useful life \cite{ramasso2014performance, zheng2017long,huang2018remaining, heimes2008recurrent}; \textit{2) failure prediction}, which aims to predict the probability that the equipment will fail within a pre-specified time window \cite{aggarwal2018two, khorasgani2018framework}. In this paper, we focus on the remaining useful life estimation (RUL) problem.


Accurate RUL estimation enables maintenance teams to confidently skip unnecessary routine maintenance tasks when the equipment is far away from its end of life. On the contrary, when the end of the equipment's life is approaching, accurate RUL estimation provides early enough warning to the maintenance departments such that they can plan their actions in advance. 

For the RUL estimation problem, there are two types of solutions: the model-based approaches and the data-driven approaches \cite{si2011remaining,zheng2017long}. Model-based methods deploy domain knowledge to handcraft degradation or failure models from a physics point of view. This type of methods are well supported by domain opinions and do not require access to a large amount of actual data \cite{zheng2017long}. However, model-based approaches are unfeasible and time-consuming when the equipment is complex. With the advancement of data collection, storage and processing techniques,  data-driven solutions which utilize sensor and operational data to estimate RUL are gaining popularity. A comprehensive review of statistical methods for the RUL estimation problem can be found at \cite{si2011remaining}. 

Recently, neural networks and deep learning have attracted a lot of research and industrial interests due to the wide application of deep learning in the areas such as image and speech understanding. Deep learning uses multiple layers of neurons to learn complex models and can be efficiently deployed with the advancement on GPUs and other software innovations \cite{zheng2016accelerating}. It is encouraging to see that deep learning has been successfully applied in several applications such as self-driving cars, natural language processing, etc. Typical deep learning models include Convolutional Neural Network (CNN),  Long Short-Term Memory (LSTM) network, Deep stacking networks, spiking networks, multilayer kernel machine, etc. These models are trained using backpropagation methods. Methods to estimate RUL from the neural network communities include a Convolutional Neural Network (CNN) based approach \cite{babu2016deep}, a deep Long Short-Term Memory method \cite{zheng2017long}, a bootstrapping based Long Short-Term Memory method \cite{liao2018uncertainty}, a deep Weilull Recurrent Neural Network approach \cite{aggarwal2018two}, a multi-task learning network \cite{aggarwal2018two}, a Hidden Markov Model (HMM) \cite{baruah2005hmms}, and neural networks with sliding window techniques \cite{wu2007neural, tian2012artificial, le2013remaining}. A detailed comparison of these neural networks in terms of RUL estimation is given in the introduction section of \cite{zheng2017long}. 

When developing data-driven solutions for RUL estimation, we often use sensor and operational time series data from multiple equipment to train RUL estimation models. In these approaches, we assume that there are correlations among the observations from the same equipment. In addition, there exist random variations across the sensor time series from multiple equipment due to random factors such as the environment around the device, the usage pattern of the device, etc. It is crucial to take in account both of the correlations within the same time series and the random variations across multiple time series. Multilayer Perceptron \cite{babu2016deep} and Convolutional Neural Network (CNN) ignore the correlations among the extracted features within each segment of any given equipment. Sequential learning techniques including HMM and LSTM utilize a sequential path from older past cells to the current one for each equipment, which implicitly take these two correlations into account. However, these models typically put strong assumptions on relationships among variables along the path. For instance, LSTM explicitly assumes that the mathematical mapping from the current sensor variables and information contained in prior sensor readings to the current RUL label is the same along the sequence. However, such assumptions might introduce biases on the RUL estimates.


To address the above concerns, we propose to address the RUL estimation challenge from the Functional Data Analysis (FDA) \cite{ramsay2006functional} perspective. FDA is a branch in statistics which concentrates on scenarios where for each subject there are multiple observations being continuously measured over time or space. Comprehensive descriptions for FDA and popular functional methods are provided in \cite{ramsay2006functional, mueller2005, wang2016functional}. FDA methods have provided valuable insights to use cases in numerous fields including Economics, Public health, Meteorology, Paleopathology, Graphology, Criminology \cite{ramsay2007applied}. When modeling the continuously-observed sensor data from the FDA perspective, for a given sensor, the time series from multiple equipment are treated as discretized realizations of an underlying continuous random process with unknown mean and covariance functions. The covariance function quantifies the correlation between the sensor data at any two time points. All the continuous random curves are viewed as random samples from the same underlying random distribution, and FDA aims to use these curve samples to infer information about the unknown mean and covariance functions. Therefore, FDA explicitly incorporates both the correlations within the same equipment and the random variations across equipment. FDA is naturally capable of handling scenarios where the relationship among variables varies over time, as a time dimension is considered. Sensor readings from different sensors are usually correlated. We proposed to use functional Multilayer Perceptron (functional MLP), a counterpart of the traditional MLP, to address this problem. In our experiments, we show the significant superiority of functional MLP for RUL estimation. Implementationally, FDA has less restrictive requirements on data than the state-of-art algorithms. FDA not only allows the number of observations and the observation times are different across equipment, but also allows the observations times vary across different sensors for the same equipment. 

The rest of the paper is organized as follows. Section \ref{sec2} presents the RUL estimation problem setting and the functional MLP approach \cite{rossi2002functional}. Section \ref{sec3} describes our experiment on a benchmark data set. Section \ref{sec4} concludes the paper. 

\section{Functional Data Analysis for RUL Estimation}
\label{sec2}
\subsection{Problem Setting}
Suppose that $N$ equipment of the same type are monitored within a pre-specified time range $\mathcal{T}$. For every equipment $i (i \in N)$, we have sensor readings from $R$ sensors at $M_i$ time points, $(T_{i,1},...,T_{i,M_i})$ such that $T_{i,j}\in \mathcal{T}$ for $j=1,...,M_i$. The number of observations and the observation times can be different across equipment, i.e, $M_i \neq M_{i^\prime}$ for $i \neq i^\prime$. The RUL for the $i$ equipment since the last observation in $\mathbf{X}_{i}$ is denoted as $Y_i$. The goal of the RUL estimation problem is to learn a mathematical mapping from $\mathbf{X}_{i}$ to $Y_i$.



When modeling from FDA point of view, for every equipment $i (i \in N)$, the $M_i$ sensor readings from a given sensor $r$ are regarded as discretized realizations of the underlying continuous curve $X_{i,r}(t)$ with $t\in \mathcal{T}$ and $r=1,...,R$. The RUL estimation problem is essentially about learning the following mapping from continuous sensor observations to the RUL label $Y_i$,

\begin{equation}
Y_i = F(X_{i,1}(t), ...., X_{i,R}(t)).\label{setting1}
\end{equation}

\begin{figure}[htbp]
\vspace{-0.12in}
\centering\includegraphics[width=5.9cm]{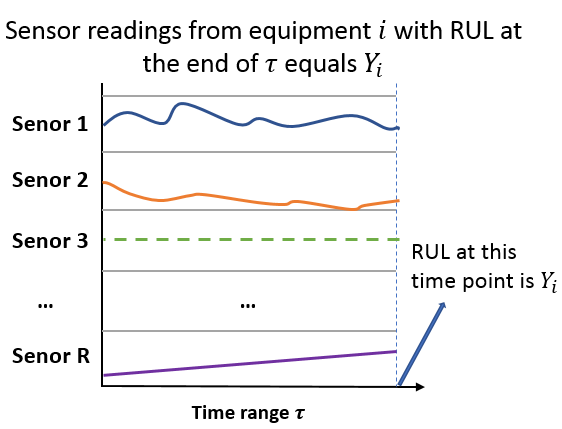}
\vspace{-0.1in}
\caption{Illustration of the problem setting for RUL estimation.}
\vspace{-0.18in}
\label{ps}
\end{figure}

The RUL estimation problem setting from the functional data perspective is demonstrated in Fig.~\ref{ps}.


\subsection{Proposed Functional Multilayer Perceptron}

Learning the mapping defined in Eq.~\eqref{setting1} is identified as the functional regression problem, where the input features are continuous random functions and the output is a numerical variable. The functional regression is considered in \cite{cardot2003spline, cai2006prediction, muller2005time, yao2005functional} where the mapping function $F(\cdotp)$ is assumed to be linear in terms of parameter functions. Functional Multilayer Perceptron (functional MLP), an extension of the classic Multilayer Perceptron to functional inputs scenarios, has been studied in \cite{rossi2005functional,  rossi2002functional,conan2002multi}. Functional MLP, an algorithm that enables non-linear learning for the functional regression problem, is capable of discovering complex relationships between the continuous sensor curves and the RUL value. Due to the complicated nature of equipment, we believe that functional MLP is more suitable for the RUL estimation problem. 

\begin{figure}[htbp]
\vspace{-0.2in}
\centering\includegraphics[width=7.75cm]{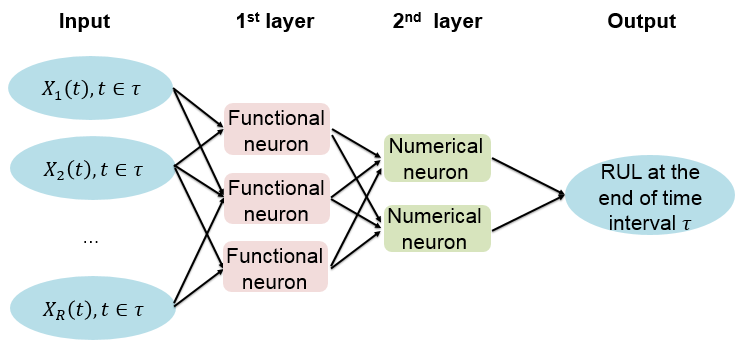}
\vspace{-0.05in}
\caption{Example of a two-layer functional MLP for the RUL estimation problem. There are three functional neurons in the first layer and two numerical neurons in the second layer.}
\vspace{-0.12in}
\label{fmlp}
\end{figure}

RUL estimation using functional MLP proceeds as follows. Functional MLP consists of a layer of functional neurons followed by multiple layers of numerical neurons as in classic MLP. Functional neurons take the continuous sensor curves as input and output numerical numbers which are then fed into the subsequent numerical layers. An example of functional MLP with three functional neurons in the first layer and two numerical neurons in the second layer is shown in Fig.~\ref{fmlp}. Analogous to numerical neurons, functional neurons consists of a linear transformation step and a non-linear transformation step. In the linear transformation step, functional neurons compute the integral of the multiplication of a specific feature curve and a weight function within time range $\mathcal{T}$, as a generalization of vector inner product in $L^2(t)$. In the non-linear transformation step, the numerical outputs in the previous step are fed into numerical activation functions.  To simplify the description for the mathematical definition of functional MLP, let's consider a two layer functional MLP where there are $K$ functional neurons on the first layer and one numerical neurons on the second layer. Mathematically, let the weight function for the $r$-th functional feature $X_{i,r}(t)$ in the $k$-th functional neuron be denoted as $V_{k,r}(\ve{\beta}_{k,r},t)$ for $k=1,...,K$ and $r=1,..,R$. The weight function $V_{k,r}(\ve{\beta}_{k,r},t)$ is assumed to be an easily computable function determined by a $Q_{k,r}$-dimensional vector $\ve{\beta}_{k,r}$. Let the activation function in the $k$-th functional neuron be denoted by $U_k(\cdotp)$, which is a numerical mapping from $\mathbb{R}$ to $\mathbb{R}$ with numerical parameters $a_k$ and $b_k$. To simplify the notation, let the concatenated unknown parameters in the weight functions across $R$ features and $K$ functional neurons be $\ve{\beta}=[\ve{\beta}_{1,1},...,\ve{\beta}_{1,R},....,\ve{\beta}_{K,1},...,\ve{\beta}_{K,R}]^T$. And the $R$ functional feature curves of the $i$-th subject is denoted as $\mathbf{X_i}=[X_{i,1}(t),..., X_{i,R}(t)]$. Then the real output of the first layer $H(\mathbf{X_i}, \ve{\beta})$ is 
\begin{equation}
H(\mathbf{X_i}, \ve{\beta})=\sum_{k=1}^{K}a_k U_k(b_k + \sum_{r=1}^{R} \int_{t\in \mathcal{T}}V_{k,r}(\ve{\beta}_{k,r},t)X_{i,r}(t)dt).\label{FN1}
\end{equation}

The choice of the weight functions $V_{k,r}(\ve{\beta}_{k,r},t)$ in Eq.~\eqref{FN1} affects functional MLP's performance, as they determine the quality of numerical features extracted by the functional neuron layer. In previous functional MLP literature \cite{rossi2005functional,  rossi2002functional,conan2002multi}, the weight functions are specified as a linear combination of B-spline functions, which are fixed functions that are not related to data. In this paper, we propose to specify data-driven weight functions by calculating the eigenfunctions from data. Specifically, the weight function of the $r$-th sensor variable in the $k$-th functional neuron $V_{k,r}(\ve{\beta}_{k,r},t)$ is a linear combination of the eigenfunctions of the $r$-th functional sensor \cite{silverman1996smoothed}.  To experimentally verify our intuition about the use of data-driven eigen function basis, we tried three different basis functions (the eigen, the B-spline and the wavelet basis functions) on a subset of data and found that eigen basis produced the best results in terms of accuracy. Before formally writing out the formula of the weight function, we briefly summarize some basics about functional principal component analysis. Functional principal component analysis plays a key role in FDA. It has been extensively investigated \cite{silverman1996smoothed} and has been utilized as a tool to create methodologies in functional linear regression, classification and clustering for functional data \cite{muller2005time}.  When modeling the repeatedly-observed sensor data from the FDA perspective, for a given sensor, the time series from multiple equipment are treated as discretized realizations of a continuous underlying random process with unknown mean and covariance functions. The covariance function quantifies the correlation between the sensor data at any two time points. Functional  Data  Analysis  (FDA) uses functional multilayer perceptron and consists of multi layers of neurons to take continuous sensors curves as input and output numerical numbers. Let all the $r$-th sensor be denoted by $X_{i,r}$ for $i=1,...,N$. The sample covariance function $G_r(s,t)$ which quantifies the correlation between the $r$-th sensor measurements at any two time points within the time domain is  
\begin{equation}
    G_r(s,t) =\frac{1}{N}\sum_{i=1}^{N} (X_{i,r}(t)-\bar{X}_r(t))(X_{i,r}(s)-\bar{X}_r(s)),
\label{cov}
\end{equation}
with $\bar{X}_r(t)=\frac{1}{N}\sum_{i=1}^N X_{i,r}(t)$. The eigenfunctions $\phi_{r,p}(t)$ are solutions of the following equation: 
\begin{equation}
    \lambda_{r,p} \phi_{r,p}(t) =\int  G_r(s,t) \phi_{r,p}(s)\,ds,
\label{cov2}
\end{equation}
where $\lambda_{r,p}$ for $p=1,...,\infty$ are the non-increasing eigenvalues that quantify the importance of the corresponding eigenfunctions. The proposed weight function $V_{k,r}(\ve{\beta}_{k,r},t)$ is specified by the first $P_r$ eigenfunctions:

\begin{equation}
\label{v_fun}
V_{k,r}(\ve{\beta}_{k,r},t) = \sum_{p=1}^{P_r} \beta_{k,r,p} \phi_{r, p}(t).
\end{equation}

\subsection{Learning and applying of Functional MLP}
Before learning and applying the proposed functional MLP, we need to first estimate the eigenfunctions $\phi_{r, p}(t)$ and determine the number of projections $P_r$ involved in the weight function in Eq.~\eqref{v_fun}. Without loss of generality, let's assume that all the subjects share a set of equally spaced observation times within window $\mathcal{T}$, which are denoted as $T_{1},...,T_{M}$. When this assumption doesn't hold, people typically use a pre-smoothing technique to first individually recover the entire underlying curves and then evaluate them at an equally spaced grid. All the observations from the $r$-th sensor can then be stored in a $N\times M$ matrix $\mathbf{X}_r$, where the $i$-th row represents the $M$ sensor measurements from the $i$-th equipment. The covariance matrix $\tilde{\mathbf{G}}_r$ which quantifies correlation between any two time points among $T_{1},...,T_{M}$ can be calculated accordingly. Let's denote the eigenvector and eigenvalues of $\tilde{\mathbf{G}}_r$ as $\tilde{\ve{\phi}}_{r,p}$ and $\tilde{\lambda}_{r,p}$ respectively, which can be computed by matrix operations. The recovered continuous function from vector $\tilde{\ve{\phi}}_{r,p}$ is denoted as $\tilde{\phi}_{r,p}(t)$. The eigenfunction and eigenvalue of the continuous covariance function $G_r(s,t)$ for $s,t\in \mathcal{T}$ are:
\begin{align} 
\label{fpca}
\hat{\phi}_{r,p}(t) &=  \sqrt{M}\tilde{\phi}_{r,p}(t) \\ 
\hat{\lambda}_{r,p} &=  \tilde{\lambda}_{r,p}.
\end{align}
As for the value of $P_r$, let the selected value for the regularly-used fraction of variance explained (FVE) for the $80\%$ cutoff be $\hat{P}_{r, \text{FVE}}$. We then propose to use $\hat{P}_r = \min \{\hat{P}_{r, \text{FVE}}, 2\}$.
By plugging in these estimates back to Eq.~\eqref{v_fun}, we get an implementable weight function $\hat{V}_{k,r}(\ve{\beta}_{k,r},t) = \sum_{p=1}^{\hat{P}_r} \beta_{k,r,p} \hat{\phi}_{r, p}(t)$.

Learning of functional MLP: Let the estimated RUL for the $i$-th data instance be $\hat{Y}_i$. The RUL estimation task is to find the optimal values for parameters in the functional MLP, including $a_k$, $b_k$, $\ve{\beta}$ in the functional layer and the relevant parameters in numerical layers, such that the mean squared error of the estimated RUL in Eq.~\eqref{obj} is minimized. 

\begin{equation}
J=\frac{1}{N}\sum_{i=1}^{N} (Y_i -\hat{Y}_i )^2.\label{obj}
\end{equation}
The optimization problem in Eq.~\eqref{obj} can be solved by gradient descent methods as described in \cite{rossi2002functional} by alternating between two steps until the stopping criterion is reached: 
\begin{itemize}[leftmargin=*]
    \item Forward propagation step: for any given parameters $a_{k}$, $b_{k}$, $\ve{\beta}$, feed the input functional features $\mathbf{X_i}=[X_{i,1}(t),..., X_{i,R}(t)]$ into the pre-specified functional neural network (an example is given in Fig.~\ref{fmlp}) to calculate the predicted RUL $\hat{Y}_{i}$. Note that $\int_{t\in \mathcal{T}}\hat{V}_{k,r}(\ve{\beta}_{k,r},t)X_{i,r}(t)dt$ in the functional neurons can be approximated by regular numerical integration methods.  We use the following approximator
    \begin{equation}
    \label{num}
    \frac{1}{M}\sum_{j=1}^{M}\hat{V}_{k,r}(\ve{\beta}_{k,r},T_{j})X_{i,r}(T_{j}).
    \end{equation}
    \item Backward propagation step: when a pre-determined stopping criteria is not satisfied, the parameters are updated by subtracting the gradient of objective function $J$ at current values of $a_{k}$, $b_{k}$, $\ve{\beta}$. Under the assumption that $\frac{\partial \hat{V}_{k,r}(\ve{\beta}_{k,r}, t)}{\partial \beta_{k,r,q}}$ exists almost everywhere for $t\in \mathcal{T}$, the gradient with respect to $\beta_{k,r,q}$ has the following formula,
    \begin{multline}
    \label{FN3}
    \frac{\partial H(\mathbf{X_i}, \ve{\beta})}{\partial \beta_{k,r,q}}=a_k U_k^\prime(b_k + \sum_{r^\prime=1}^{R} \int \hat{V}_{k,r^\prime}(\ve{\beta}_{k,r^\prime}, t)X_{i,r^\prime}(t)dt)\\ 
    \times \int_{t\in \mathcal{T}}\frac{\partial \hat{V}_{k,r}(\ve{\beta}_{k,r}, t)}{\partial \beta_{k,r,q}}X_{i,r}(t)dt.
    \end{multline}
    Numerical integrations in Eq.~\eqref{FN3} are approximated in the same fashion as Eq.~\eqref{num}. The updated parameters are then used in the following forward propagation step. 
\end{itemize}
Let's denote the learned model in the training phase as $\hat{F}_{fmlp}(\mathbf{X_i})$, where $F_{fmlp}(\mathbf{X_i})$ is a non-linear mapping corresponding to the pre-specified functional MLP structure whose optimal parameter are learned by the gradient descent method described above.

Application of functional MLP: For any new equipment which are monitored for a time window of length $\mathcal{T}$, we can deploy the learned model to estimate the RUL. Specifically, let's assume that the $r$-th sensor variables are measured at the same time grids as the equipment in the training set and denote the observed sensor values for the $r$-th feature as vector $\mathbf{X}_{new, r}=[X_{new, r, 1},...,X_{new, r, M}]^T$. If this assumption doesn't hold, we can achieve such data by first recovering the entire curve and then evaluating at these times. The estimated RUL for this new equipment is the output of $\hat{F}_{fmlp}(\mathbf{X}_{new, 1},...,\mathbf{X}_{new, R})$, which can be calculated by supplying the feature vectors into the learned functional neurons, followed by feeding the output of function neuron into the subsequent numerical neurons. The output from the learned functional layer is 
\begin{equation}
\sum_{k=1}^{K}\hat{a}_k U_k(\hat{b}_k + \sum_{r=1}^{R} \sum_{j=1}^{M}\frac{1}{M}\hat{V}_{k,r}(\hat{\ve{\beta}}_{k,r},T_{j})X_{new,r,j}).\label{apply}
\end{equation}

\section{Experiments on C-MAPSS Data Set}
\label{sec3}
In this section, we apply functional MLP (`FMLP') to conduct RUL estimation task for a widely-used benchmark data set called NASA C-MAPSS (Commercial Modular Aero-Propulsion System Simulation) data \cite{saxena2008phm08}. We compare the performance of functional MLP with a variety of state-of-the-art deep learning approaches, including the Convolutional Neural Network model (`CNN') in \cite{babu2016deep}, the Deep Weibull network (`DW-RNN') and the multi-task learning network (`MTL-RNN') in \cite{aggarwal2018two}, the Long Short-Term Memory method (`LSTM') \cite{zheng2017long}, and the bootstrapping based Long Short-Term Memory method (`LSTMBS') \cite{liao2018uncertainty}. As shown by the experimental results, the proposed functional MLP approach significantly outperforms all these alternative methods.

\subsection{Background}
C-MAPSS data set contains of simulated sensor readings, operating conditions for a group of turbofan engines as they running until some critical failures happen. For each turbofan engine, there are three time series data which quantify the different operating conditions over time and 21 sensor time series being recorded within the time window considered. The schema of the simulator and the physical meanings of the sensor and operating condition variables are discussed in detail in \cite{saxena2008damage}. The goal of the RUL estimation problem is to use the sensor and operating data trajectories from the start of measurements till a certain time $t$ to estimate the RUL of the engine at $t$. It is very difficult to estimate RUL using these sensor data with different operating conditions. 

To mimic the real practice where there are a mix of operating conditions on the engines over time and the engines might fail due to different root causes, scenarios with different numbers of operating conditions and fault conditions are simulated by the simulator \cite{saxena2008damage}. Specifically, C-MAPSS data set consists of four data subsets. Tab.~\ref{bg} summarizes the number of operating conditions and fault modes in each subset. The individual subsets are divided into training and testing sets. The training sets contain run-to-failure data where engines are fully observed from an initial healthy state to a failure state. The testing sets consist of prior-to-failure data where engines are observed until a certain time before failure. True RUL labels associated with engines in the testing sets are provided for evaluation purpose. Estimating RUL for this data set is a hard problem, because different operating conditions are mixed together, and from raw sensor data, it is not straightforward to see clear trends along the degradation of engine performance. Thus, many complex data-driven models, such as Convolutional Neural Network, Long Short-Term Memory networks, were applied on this task \cite{zheng2017long}. However, existing algorithms have theoretical flaws as discussed in Section \ref{sec1}. In the remainder of this section, we present in detail how we achieve significantly improved RUL estimations using functional MLP and explain why our model works so well.


\begin{table}[htbp]
\vspace{-0.08in}
\caption{Summary of the subsets in C-MAPSS data set}
\vspace{-0.15in}
\begin{center}
\begin{tabular}{c|cccc}
\hline
\hline
\textbf{}& \textbf{FD001}&   \textbf{FD002}&  \textbf{FD003}& \textbf{FD004}\\
\hline
$\#$ of engines in training& 100 & 260 & 100  & 249  \\
$\#$ of engines in testing& 100 & 259 & 100 &  248 \\
$\#$ of operating conditions& 1 & 6 & 1 &   6 \\
$\#$ of fault modes & 1 & 1 & 2 &  2 \\
\hline
\hline
\end{tabular}
\vspace{-0.17in}
\label{bg}
\end{center}
\end{table}

\subsection{Removing the effect of operating conditions}
 In FD002 and FD004, there are six operating conditions reflected in three numerical operating condition variables. The existence of multiple operating conditions makes it inappropriate to estimate RUL based on the sensor time series alone. Operating conditions clearly affect the sensor readings as well as RUL. It is essential to take the operating conditions into account in the modeling process.

 
 Some of the work on this data set \cite{babu2016deep, zheng2017long} cluster observations at different time cycles across all the engines in each subset into six distinct clusters. The three continuous operating condition variables are then replaced by three binary variables which indicate the cluster label. The new operating condition variables together with the sensor variables are fed into CNN or LSTM to perform the RUL estimation task. 
 
\begin{figure*}[htbp]
	\centering
	\begin{subfigure}[t]{7.05in}
		\centering
        \caption{Raw sensor data for one randomly selected engine in the training set of FD002 (6 operating conditions)}\label{raw}
		\includegraphics[width=14.75cm,height=3.25cm]{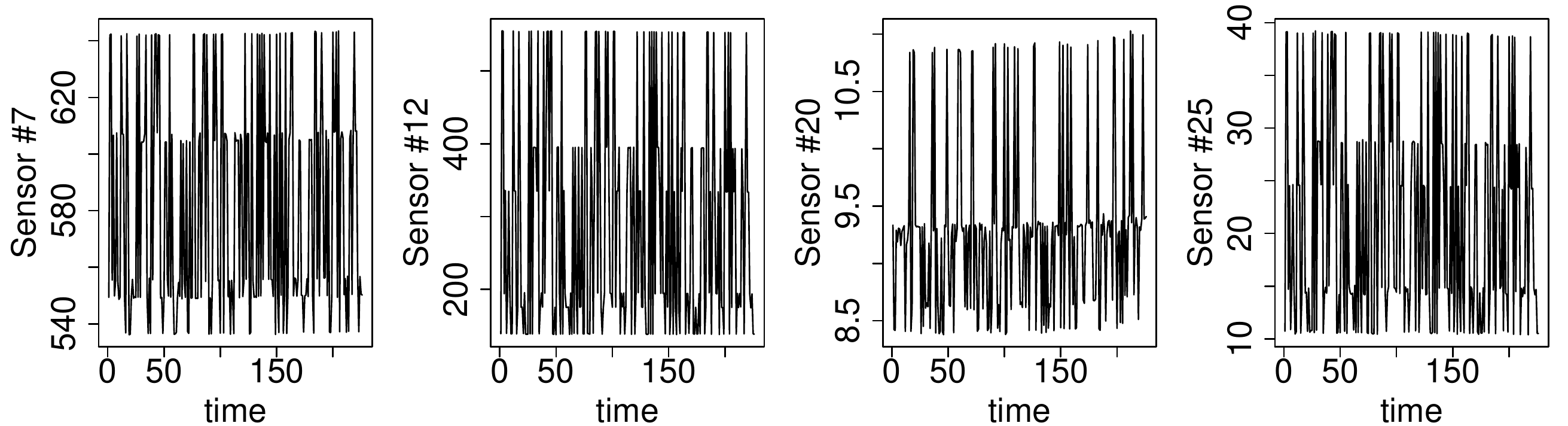}		
	\end{subfigure}
    \par\bigskip
	\begin{subfigure}[t]{7.05in}
		\centering
        \caption{Normalized sensor data after removing the effect of operating conditions}\label{normalized}
		\includegraphics[width=14.75cm,height=3.25cm]{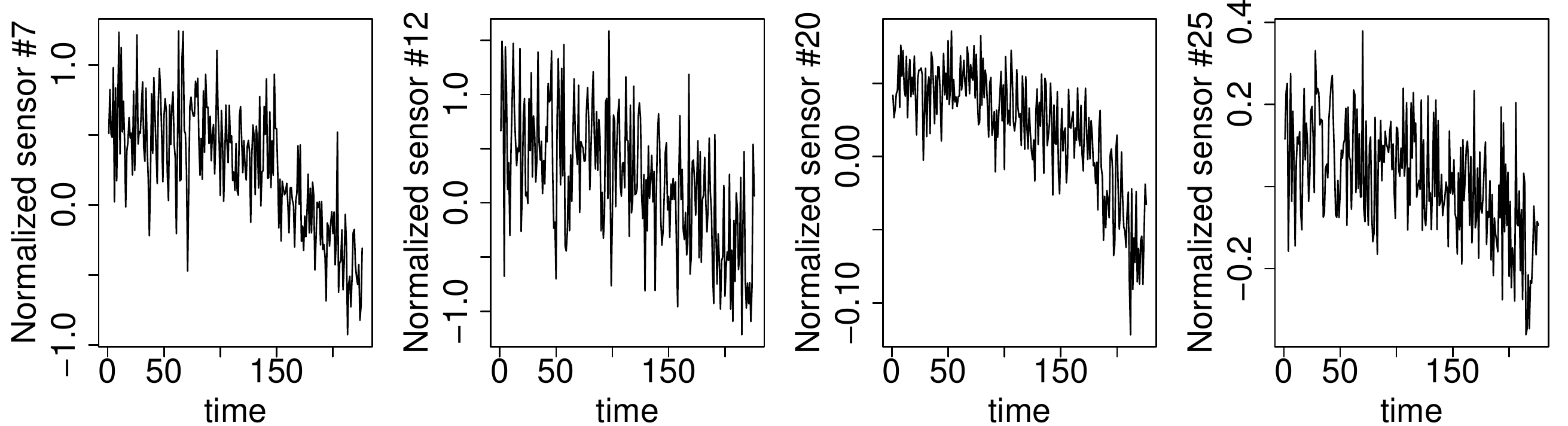}		
	\end{subfigure}
	\vspace{-0.05in}
	\caption{Removing the effect of operating conditions on sensor data}\label{normalization}
\end{figure*}

 We consider an alternative approach to handle variability due to the operating conditions.  We propose to remove the effects of operating conditions on the sensor data through the following procedure. For each of the 21 sensor data, we use the data from each training set to train a regression model which maps from the operating condition variables to the sensor variable. The achieved regression model enables us to estimate the would-be sensor data given any operating condition. We then calculate the normalized sensor data by deducting the would-be sensor data from the raw sensor readings. In our experiment, we train the regression model using Multilayer Perceptron. The normalization process is demonstrated by Fig.~\ref{normalization}. The raw sensor time series of a randomly selected engine in the training set of FD002 are plotted in Fig.~\ref{raw} and the normalized sensor data are visualized in Fig.~\ref{normalized}. Fig.~\ref{normalization} shows that the hidden trends in the sensor time series are discovered by our normalization step. In later experiments, we use the normalized sensor data to conduct the RUL estimation task. The same regression-based normalization procedure is also considered in \cite{wang2018maintenance}.

\subsection{Data preparation}
\begin{figure}[htbp]
\centering\includegraphics[width=8.05cm]{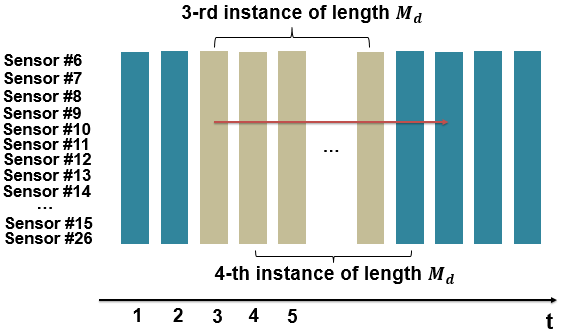}
\caption{Illustration of the one-step window sliding technique using one equipment data traojectories. }
\label{ws}
\end{figure}

\begin{figure}[htbp]
	\centering
	\begin{subfigure}[t]{1.55in}
		\centering
        \caption{Remaining useful life label}\label{RUL_lbl}
		\includegraphics[width=40mm]{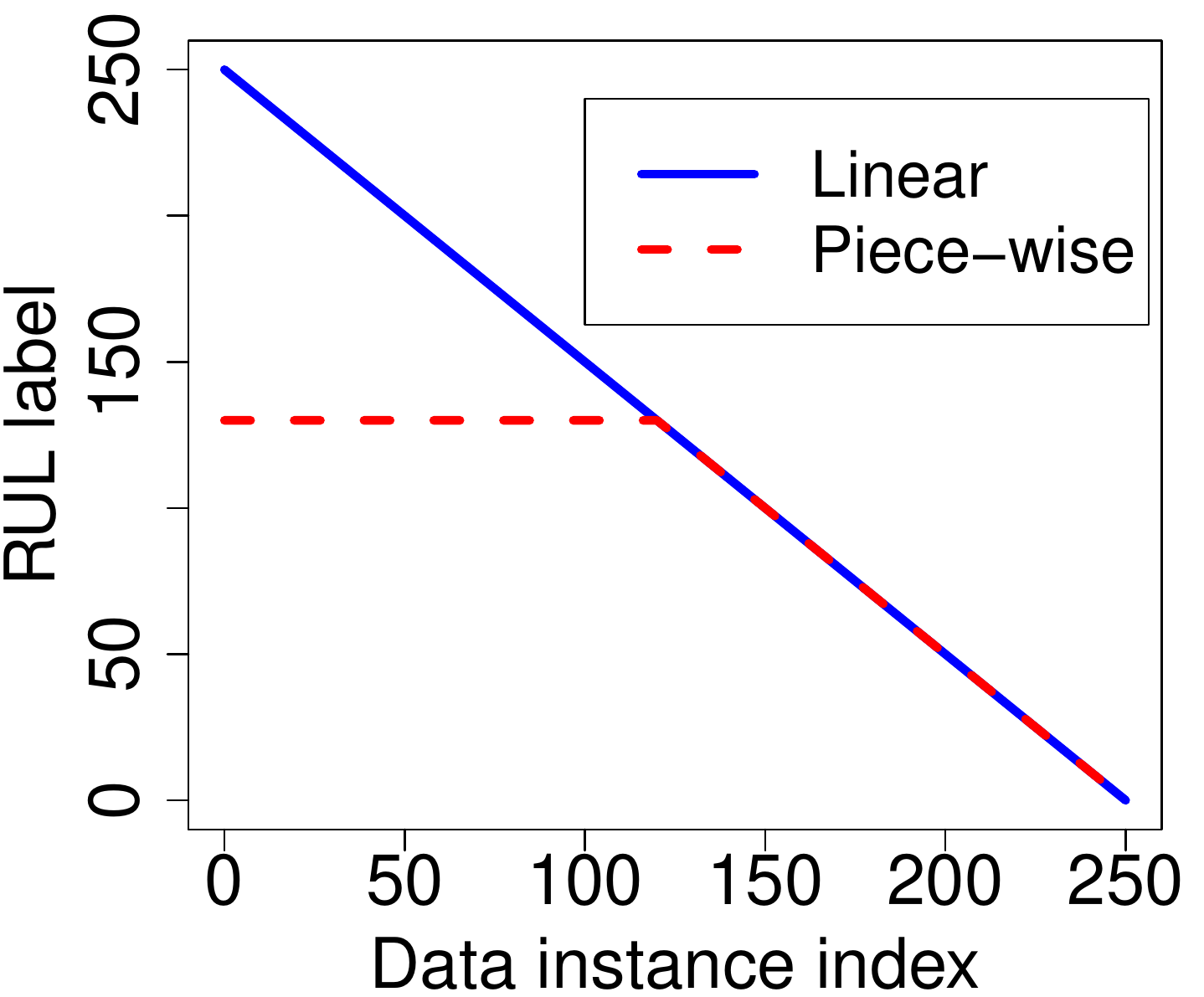} 		
	\end{subfigure}
	\quad
	\begin{subfigure}[t]{1.55in}
		\centering
        \caption{Evaluation metrics}\label{metric}
		\includegraphics[width=40mm]{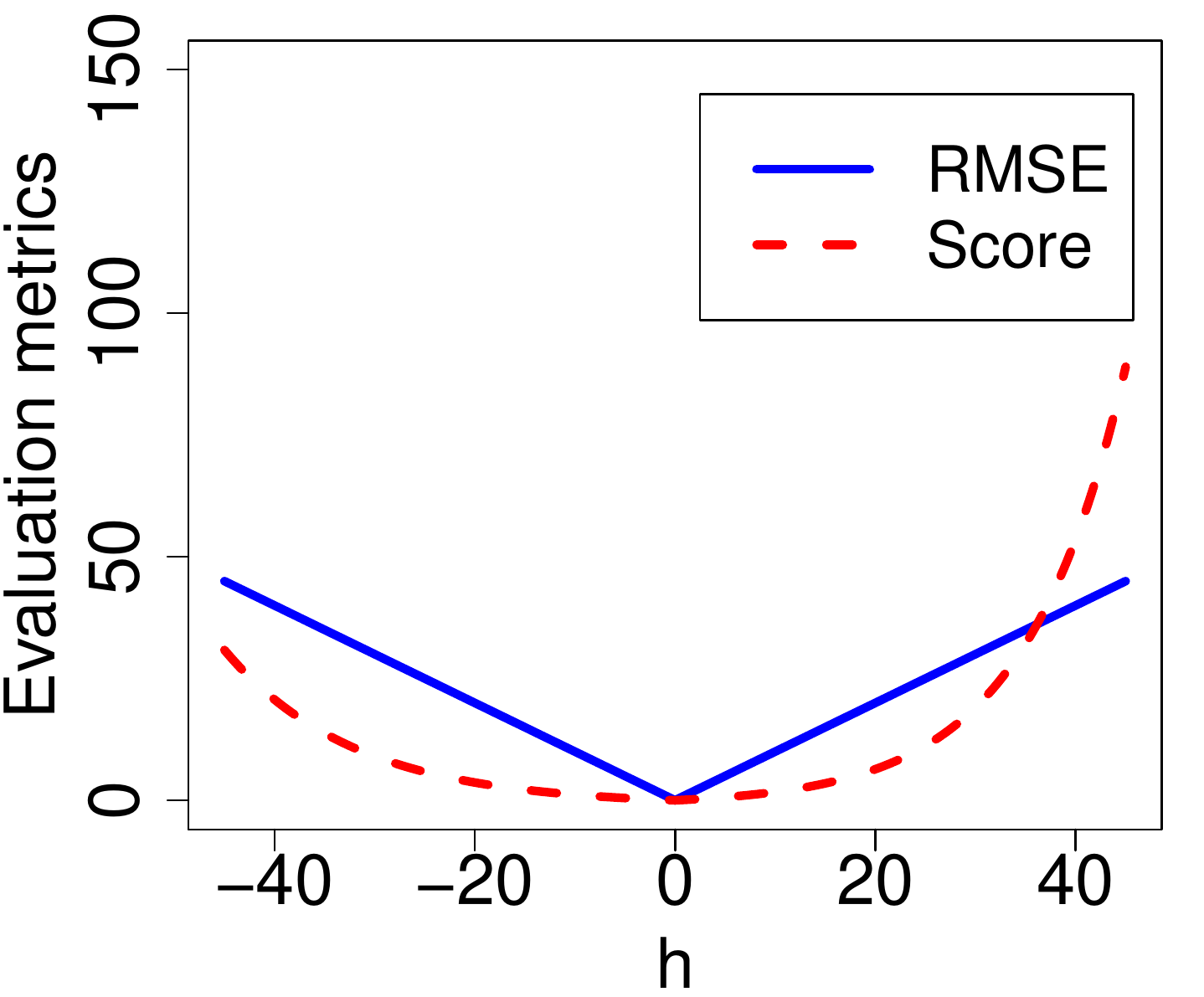} 		
	\end{subfigure}
	\caption{Left: Remaining useful life label for a given engine, the blue line represents the piece-wise RUL label; the red line represents the piece-wise RUL label capped at 130. Right: Visualizations of two evaluation metrics (i.e., RMSE and score function) as a function of $h$.}
\end{figure}

Under the typical functional regression setting introduced in Section \ref{sec2}, each sample in the RUL estimation problem consists of continuous sensor readings within a given time range and an associated RUL label to indicate the time length from the present to the end of the subject's life. Different subjects represent different engines. However, in C-MAPSS data set, the engines in the training and testing sets are observed for different number of time cycles. Moreover, the full sensor data trajectories in the testing sets are blinded for a variety of periods, therefore the true RUL labels are distributed variously. To handle this phenomenon, we propose to use the window sliding technique used in \cite{wu2007neural, tian2012artificial, le2013remaining}. Let's denote the number of sensor measurements for a single engine in data subset $d$ as $\mathcal{M}_d$ for $d=1,...,4$. The values for 
$\mathcal{M}_1, \mathcal{M}_2, \mathcal{M}_3, \mathcal{M}_4$ are $31, 21, 38, 19$ respectively. The functional inputs and RUL labels are generated as follows. For the $d$-th subset, trajectories corresponding to each engine in the training and testing data sets are cut into multiple data instances of length $\mathcal{M}_d$. For instance, the first engine in the training set of FD001 fails at the 144th cycle. A total of 114 training data instances are generated from this engine, with the $c$-th data instance being the sensor measurements between time cycle  $c$ and $c+\mathcal{M}_d-1$. This functional inputs creation step is demonstrated by Fig.~\ref{ws}. There are two ways to specify the RUL labels for the 114 data instances of this engine: one is called the linear RUL labeling approach, which assumes that equipment performance decreases linearly along with time, 
\begin{equation}
    \text{RUL}_{c,linear} = \# \text{of data instances} - c. \label{rul_l}
\end{equation}
The other one is called the piece-wise labeling approach, which assumes that the degradation in the performance is negligible at the beginning period and it starts to degrade linearly at some point $T$,
\begin{equation}
    \text{RUL}_{c, piecewise} = \min\{T,  \text{RUL}_{c,linear}\}.\label{rul_ps}
\end{equation}
These two RUL labeling techniques are visualized in Fig.~\ref{RUL_lbl} using an engine with 250 functional data instances as an example. In Fig.~\ref{RUL_lbl}, the linear labeling approach is indicated by the blue line and the piece-wise labeling approach is represented by the red line. The piece-wise labeling approach with $T=130$ is widely utilized in relevant literature \cite{babu2016deep,zheng2017long}. To make the experimental results comparable, we use this RUL labeling approach in this paper.

\subsection{Implementation of functional MLP}

Based on the achieved data from the previous subsection, details of our experiments are summarized below. 

\textit{1) Data normalization:} For each sensor, the raw sensors measurements are normalized with the Min-Max normalization in \cite{zheng2017long} such that the sensor data are scaled to the range of $[0,1]$. Mathematically, let the $r$-th sensor across all time points and all engines are stored in a vector $\ve{O}_r$. The normalized sensors $\ve{\tilde{O}}_r$ are then calculated by 
\begin{equation}
    \ve{\tilde{O}}_r=\frac{\ve{O}_r-\min \ve{O}_r }{\max \ve{O}_r -\min \ve{O}_r },
\end{equation}
where $\min \ve{O}_r$ and $\max \ve{O}_r$ are respectively the minimum and maximum value in vector $\ve{O}_r$. 


\textit{2) Architecture of functional MLP:} When analyzing the C-MAPSS data set, we use a two-layered functional MLP model with four functional neurons (i.e., K=4) on the first layer and two numerical neurons on the second layer. The activation function on both layers are the standard logistic function, i.e., $U_{k}(u) = \frac{1}{1+e^{-u}}$. The weight functions used $V_{k,r}(\ve{\beta}_{k,r},t)$ have been discussed in Section \ref{sec2}. 

\begin{figure*}[htpb]
\centering
\begin{minipage}[t]{0.86\linewidth}
    \includegraphics[width=155.0mm]{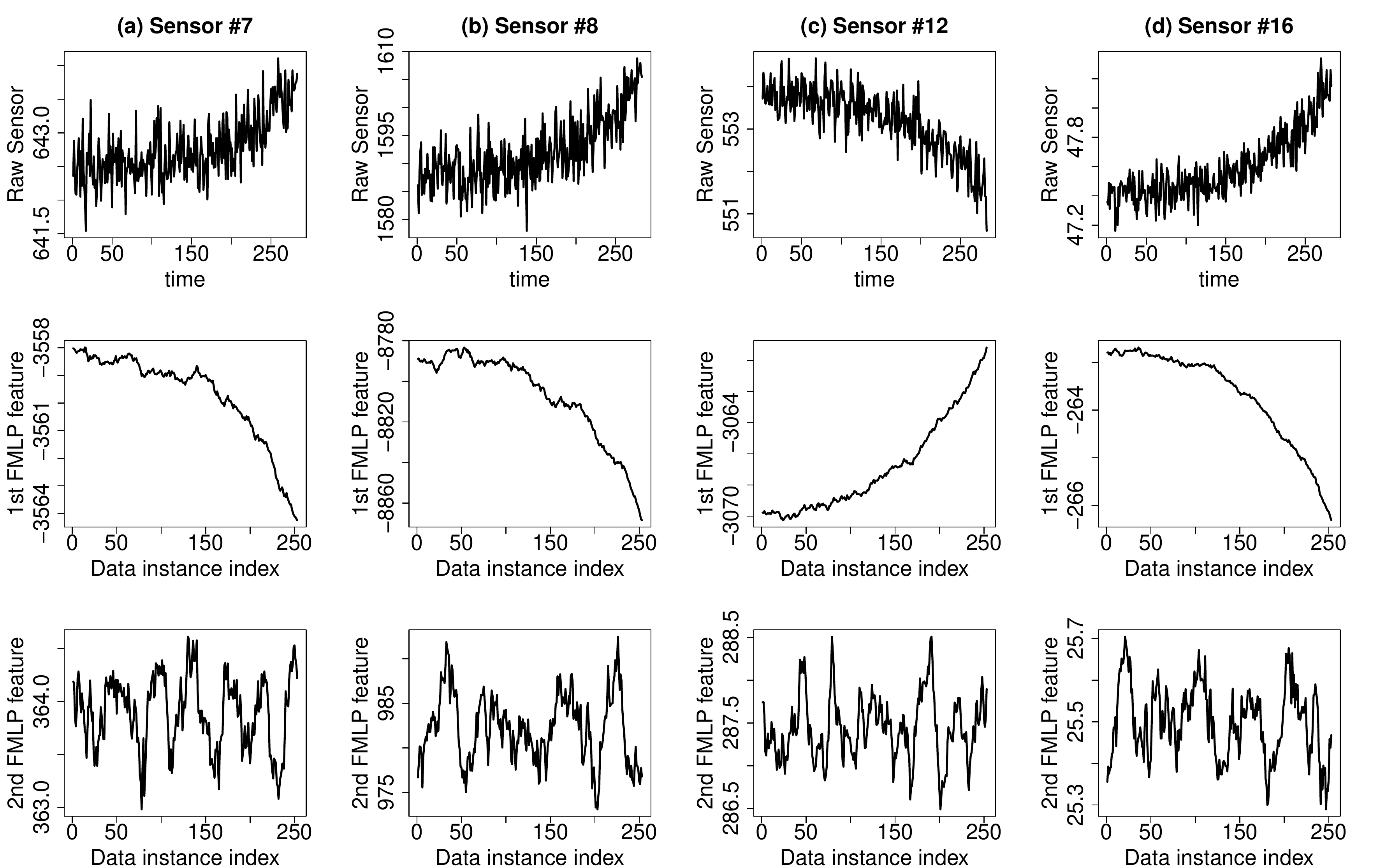}
		\caption{Visualizations of the raw sensor and the corresponding extracted features from the first layer of functional MLP for a randomly selected engine in the traning data set of FD001.}\label{feature}
	\parbox{6.5cm}{\small \hspace{1.5cm} }
\end{minipage}
\vspace{-0.3in}
\end{figure*}

The extracted features by the functional neuron specified above are visualized in Fig.~\ref{feature}. For a randomly selected engine in the training data set of FD001, the raw sensor time series of Sensor $\#7, \#8, \#12, \#16$ are provided in the upper panel of Fig.~\ref{feature}. The two extracted features along the generated data instance sequences by our specified functional neurons are presented in the second and third rows. Given Fig.~\ref{feature}, it can be seen that the extracted features from functional neurons have a clearer correlation with the performance (in this case, RUL labels) of the equipment, compared to the raw sensor data. This visually explains the exceptional performance of our functional MLP on the RUL estimation challenge.

\textit{3) Evaluation metrics:} We evaluate the performance of functional MLP with the same evaluation strategy used in \cite{babu2016deep,zheng2017long}. For each engine in the test data sets, we count the estimated RUL from the last data instance when calculating the overall estimation accuracy. This is driven by practical use cases where people often aim to estimate the RUL from the present. Suppose that there are $N$ subjects, and the true RUL since the last observation of engine $i$ is $\text{RUL}_{i, true}$ and the estimated RUL is $\text{RUL}_{i, est}$. Let $h_i=\text{RUL}_{i, est}-\text{RUL}_{i, true}$ be the estimation error for the $i$-th engine. The root mean squared error (RMSE) calculated from the $N$ engines is defined as
\begin{equation}
    \text{RMSE} = \sqrt{\frac{1}{N}\sum_{i=1}^N h_i^2}.
\end{equation}
A score-based evaluation function is 
\begin{equation}
    \text{Score} = \left \{
  \begin{aligned}
    &\sum_{i=1}^N (e^{-\frac{h_i}{13}}-1), && h_i < 0 \\
    &\sum_{i=1}^N (e^{\frac{h_i}{10}}-1), && h_i \geq 0
  \end{aligned} \right.
\end{equation}
The shapes of the two evaluation functions in terms of a single error term $h$ are visualized in Fig.~\ref{metric}. The RMSE metric is represented by the blue line and the score function is shown by the red line. It can be seen that the  RMSE metric is symmetric with respect to 0, i.e., it penalizes underestimates and overestimates of the same magnitude equally. On the contrary, the score metric tends to penalize more on scenarios when the estimated RUL is larger than the true RUL. Note that smaller values indicate better accuracies, according to the definitions of these two metrics. 


\textit{4) Results:}
The RMSE and score metrics of our functional MLP together with results from previous literature are summarized in Tables \ref{tab1} and \ref{tab2}. For all the four subsets, functional MLP significantly outperforms the baseline methods in both metrics. The improvement over LSTM are calculated by 
\begin{equation}
    \text{IMP}=1-\frac{\text{Metric of FMLP}}{\text{Metric of LSTM}}.\label{IMP1}
\end{equation}
As shown by the results, functional MLP achieves notable improvements over LSTM \cite{zheng2017long} on both the RMSE and score metrics. Specifically, the average improvement over four subsets on RMSE is $26.89\%$ and the average improvement on the score metric is $70.54\%$. 

\begin{figure}[htbp]
	\centering
	\begin{subfigure}[t]{1.55in}
		\centering
        \caption{FD001}\label{rul1}
        \vspace{-0.1in}
		\includegraphics[width=35mm, height=28mm]{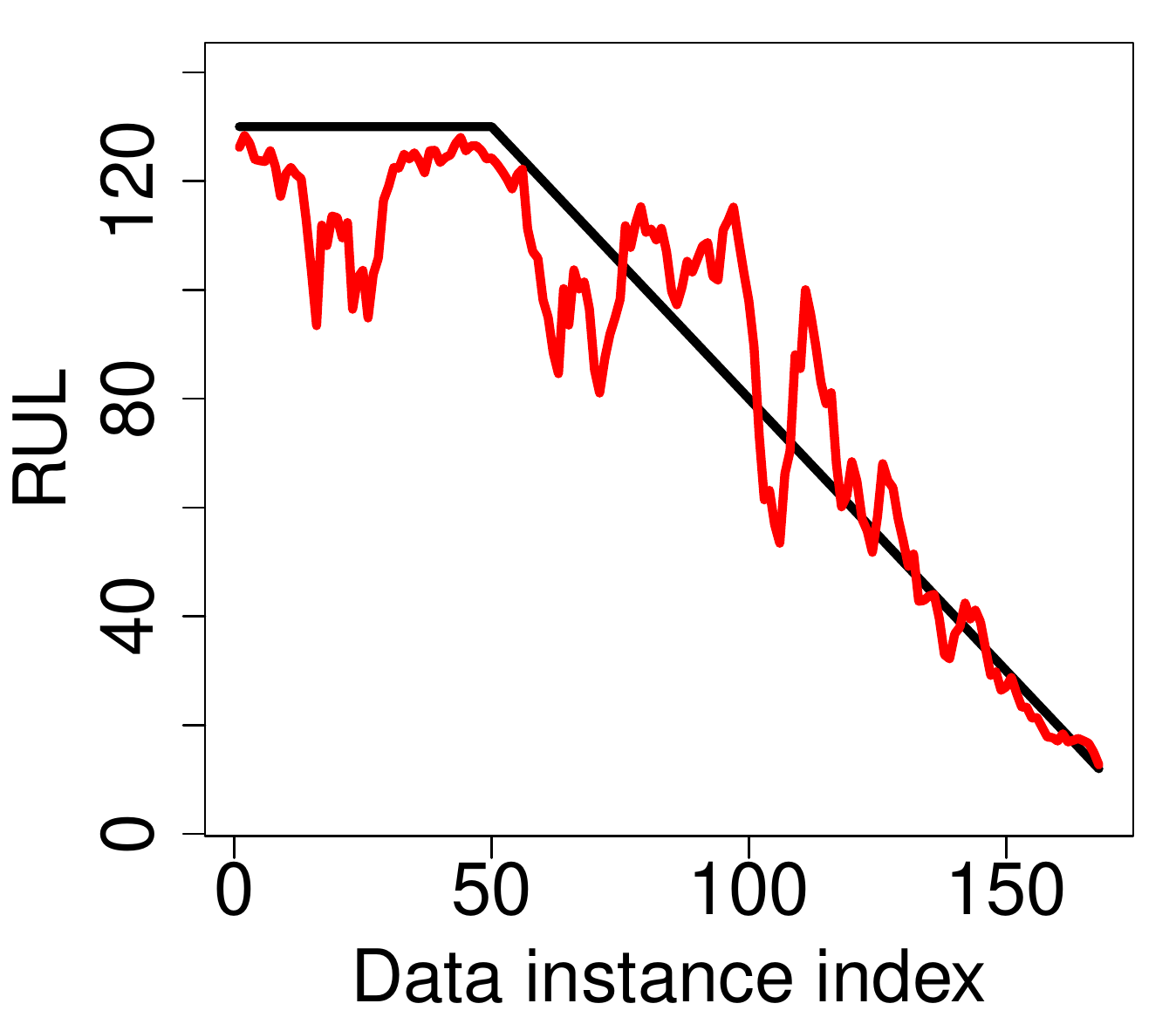} 		
	\end{subfigure}
	\quad
	\begin{subfigure}[t]{1.55in}
		\centering
        \caption{FD002}\label{rul2}
        \vspace{-0.1in}
		\includegraphics[width=35mm, height=28mm]{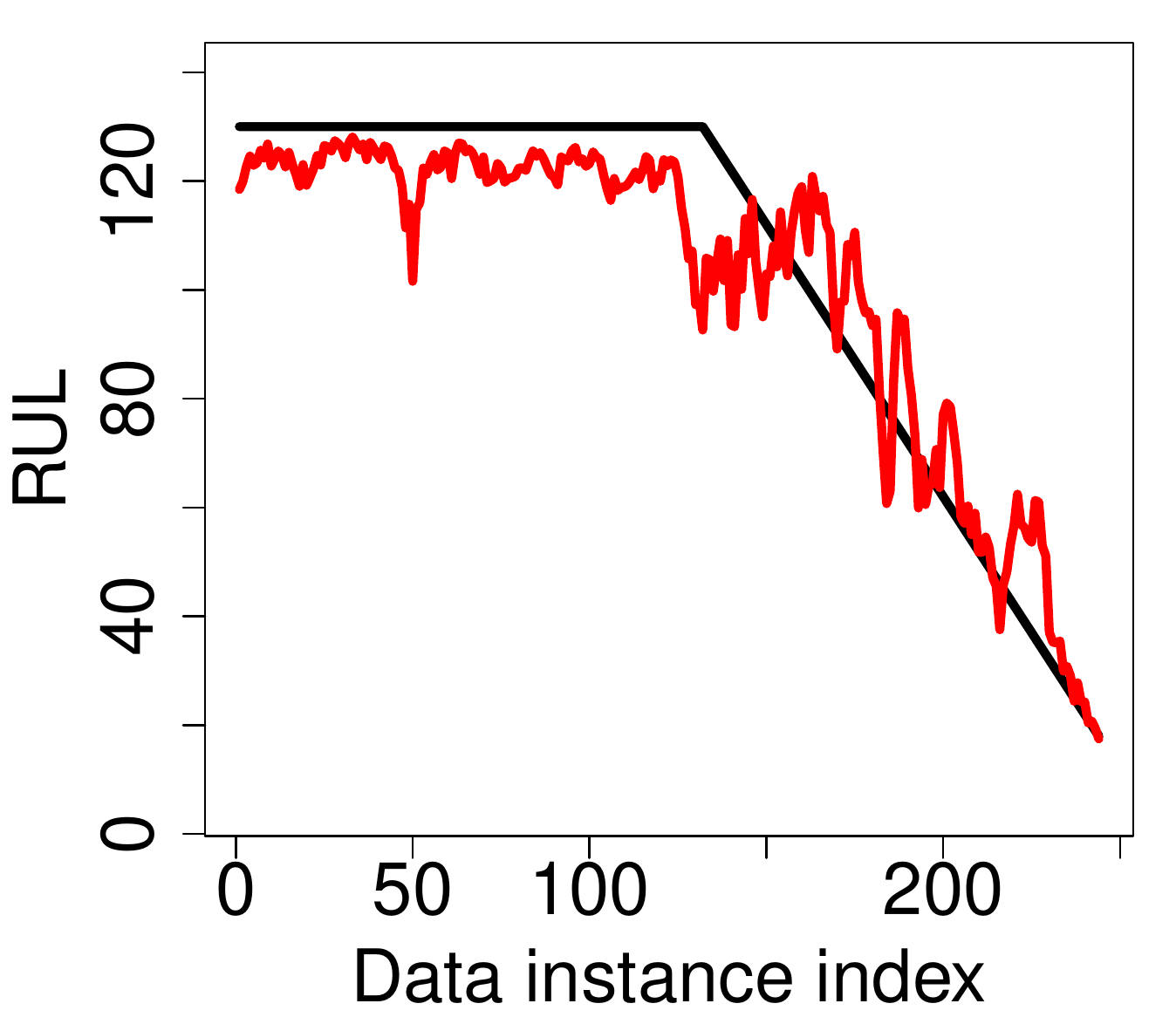} 		
	\end{subfigure}
    \par \bigskip
	\begin{subfigure}[t]{1.55in}
		\centering
        \caption{FD003}\label{rul3}
        \vspace{-0.1in}
		\includegraphics[width=35mm, height=28mm]{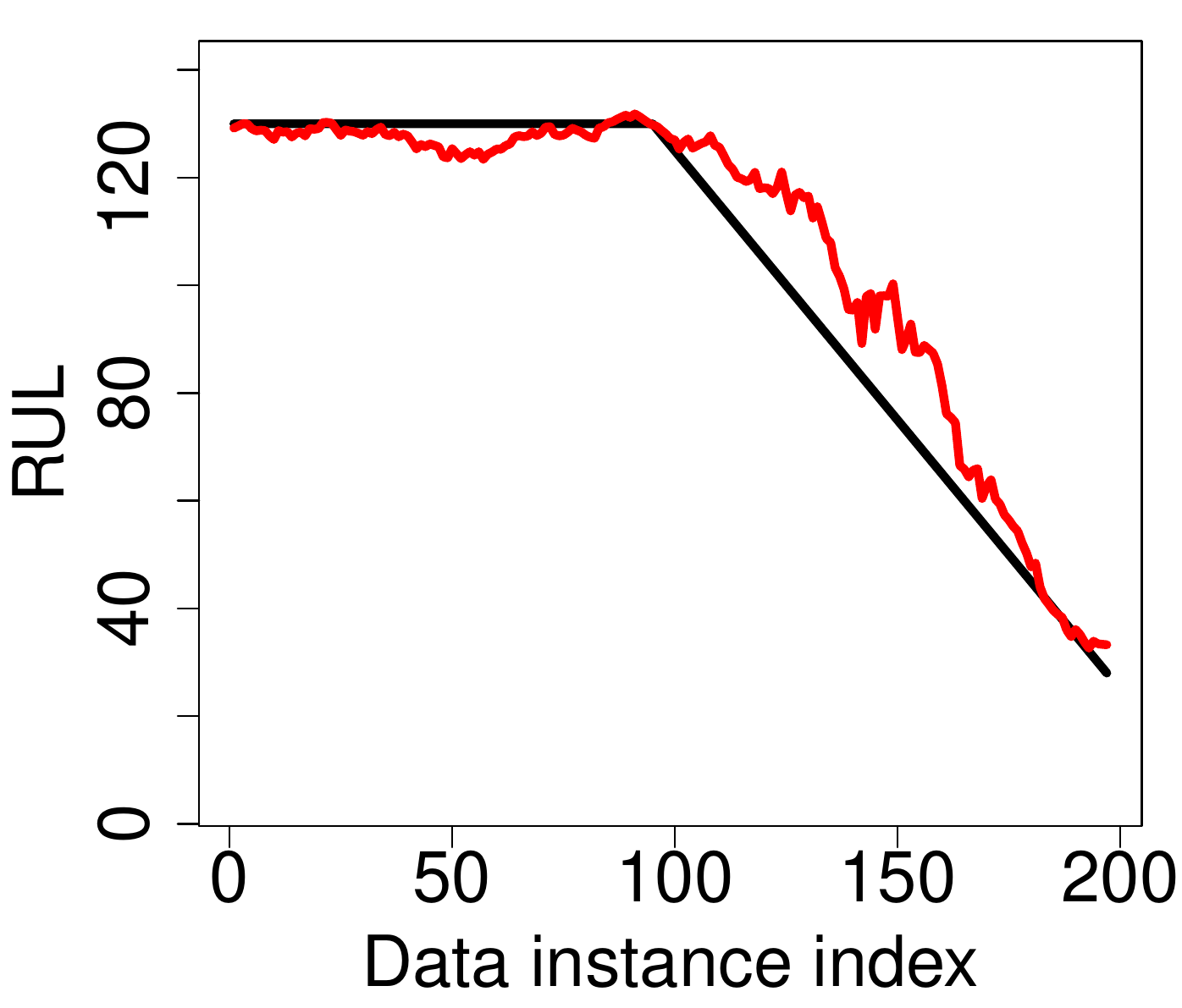} 		
	\end{subfigure}
	\quad
	\begin{subfigure}[t]{1.55in}
		\centering
        \caption{FD004}\label{rul4}
        \vspace{-0.1in}
		\includegraphics[width=35mm, height=28mm]{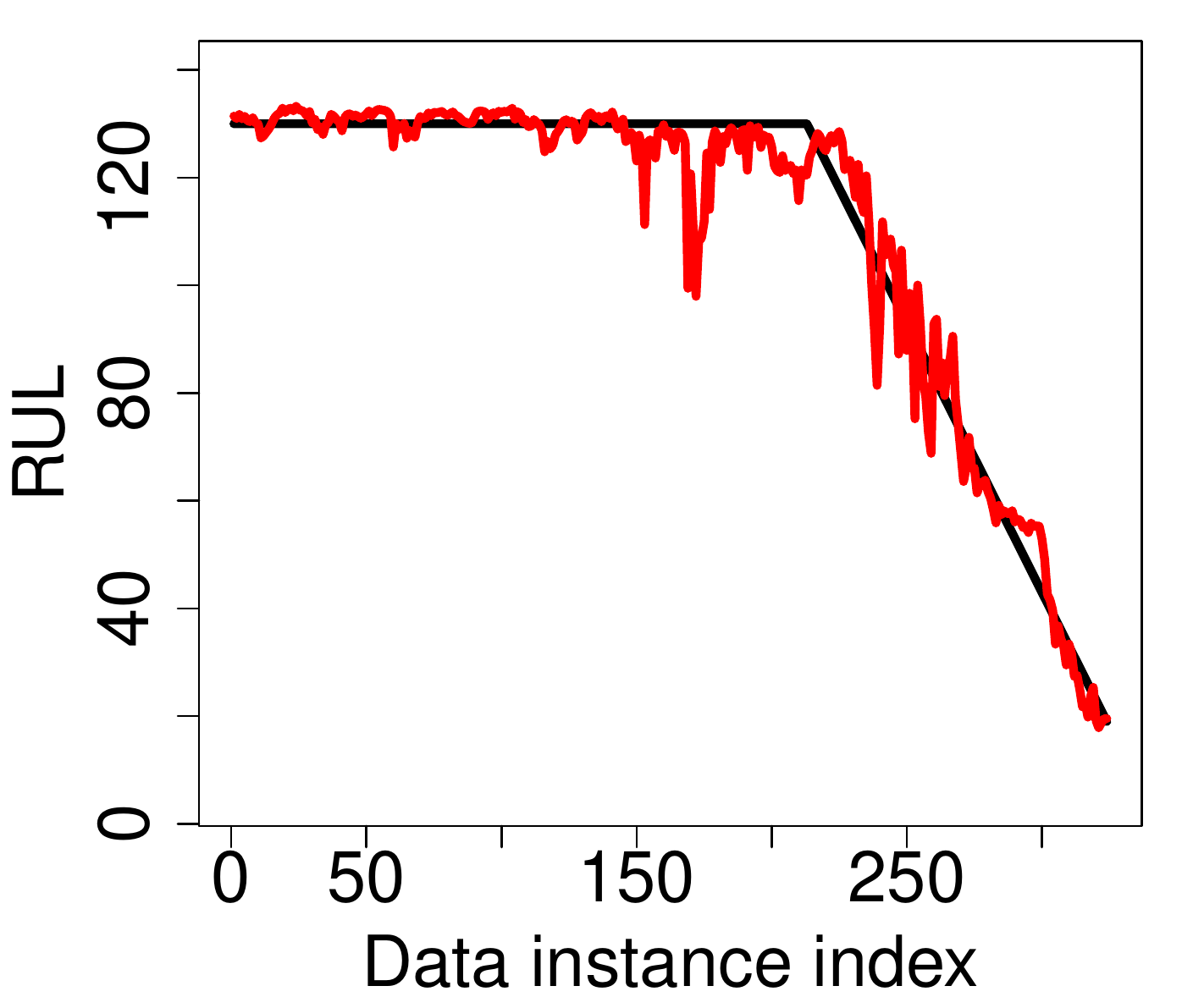} 		
	\end{subfigure}
	\caption{The true and the estimated RUL from functional MLP for a randomly selected engine in the testing sets. In each plot, the black curve is the true piece-wise RUL capped at 130 and the red line represents the estimated RUL from FMLP.}\label{rul_curve}
	\vspace{-0.2in}
\end{figure}

To examine the RUL estimation performance along the entire lifespan, we compare the true (black in Fig.~\ref{rul_curve}) and estimated RUL (red in Fig.~\ref{rul_curve}) for a randomly choose one engine in each of the four testing data sets. Generally, the plots in Fig.~\ref{rul_curve} demonstrate that the functional MLP is capable of making RUL estimation somewhat accurately. Another observation is that the estimated RUL tends to be closer to the true piece-wise RUL when it is close to the end of life.

\begin{table}[htbp]
\caption{RMSE comparison on C-MAPSS data and improvement (`IMP') of functional MLP over LSTM \cite{zheng2017long}}
\begin{center}
 \vspace{-0.1in}
\begin{tabular}{c|cccc}
\hline
\hline
\textbf{Model}& \textbf{FD001}&   \textbf{FD002}&  \textbf{FD003}& \textbf{FD004}\\
\hline
MLP\cite{babu2016deep}& 37.56 & 80.03 & 37.39  & 77.37  \\
SVR\cite{babu2016deep} & 20.96 & 42.00 & 21.05 &  45.35 \\
RVR\cite{babu2016deep} & 23.80 & 31.30& 22.37 &   34.34 \\
CNN\cite{babu2016deep} & 18.45 & 30.29 & 19.82 &   29.16 \\
DW-RNN\cite{aggarwal2018two} & 22.52 & 25.90 & 18.75 &   24.44 \\
MTL-RNN\cite{aggarwal2018two} & 21.47 & 25.78 & 17.98 &   22.82 \\
LSTMBS\cite{liao2018uncertainty} & 14.89 & 26.86 & 15.11 &   27.11 \\
LSTM\cite{zheng2017long} & 16.14 & 24.49& 16.18 &   28.17 \\
FMLP& \textbf{13.36} & \textbf{16.62} & \textbf{12.74}  & \textbf{17.76}  \\
\hline
\hline
IMP& $17.22\%$ & $32.14\%$ & $21.26\%$  & $36.95\%$  \\
\hline
\hline
\end{tabular}
\label{tab1}
\vspace{-0.1in}
\end{center}
\end{table}

\begin{table}[htbp]
\caption{Score comparison on C-MAPSS data and improvement (`IMP') of functional MLP over LSTM \cite{zheng2017long}}
\vspace{-0.1in}
\begin{center}
\begin{tabular}{c|cccc}
\hline
\hline
\textbf{Model}& \textbf{FD001}&   \textbf{FD002}&  \textbf{FD003}& \textbf{FD004}\\
\hline
MLP\cite{babu2016deep}& $1.8\times 10^4$ & $7.8\times 10^6$ & $1.7\times 10^4$ & $5.6\times 10^6$   \\
SVR\cite{babu2016deep} & $1.4\times 10^3$ & $5.9\times 10^5$ & $1.6\times 10^3$ &  $3.7\times 10^5$  \\
RVR\cite{babu2016deep} & $1.5\times 10^3$ & $1.7\times 10^4$ & $1.4\times 10^3$ &   $2.7\times 10^4$\\
CNN\cite{babu2016deep} & $1.3\times 10^3$ & $1.4\times 10^4$ & $1.6\times 10^3$ &   $7.9\times 10^3$ \\
LSTMBS\cite{liao2018uncertainty} & $4.8\times 10^2$ & $8.0\times 10^3$ & $4.9\times 10^2$ &   $5.2\times 10^3$ \\
LSTM\cite{zheng2017long} & $3.4\times 10^2$ & $4.5\times 10^3$ & $8.5\times 10^2$ &   $5.6\times 10^3$ \\
FMLP& $\mathbf{2.0\times 10^2}$ & $\mathbf{9.0\times 10^2}$ & $\mathbf{1.8\times 10^2}$  & $\mathbf{1.0\times 10^3}$  \\
\hline
\hline
IMP& $41.18\%$ & $80.00\%$ & $78.82\%$  & $82.14\%$ \\
\hline
\hline
\end{tabular}
\label{tab2}
\vspace{-0.1in}
\end{center}
\end{table}

\section{Conclusions and Discussion}
\label{sec4}
With the advancement of data-driven techniques for Prognostic and Health Management (PHM), there is a growing demand from researchers and practitioners to get accurate insights about the Remaining Useful Life (RUL) of an equipment or a component. We proposed a new perspective to address the RUL estimation challenge. We discussed the problem setting, data preparation, model construction processes of the proposed functional Multilayer Perceptron (functional MLP) model. We provided visual explanations regarding the benefits of estimating RUL with functional MLP. Our experimental results on the well-known benchmark data set called NASA C-MAPSS data demonstrated that our functional MLP significantly outperforms alternative data-driven methods.





%
\bibliographystyle{IEEEtran}
\bibliography{fanova}

\end{document}